# LeafLife: An Explainable Deep Learning Framework with Robustness for Grape Leaf Disease Recognition

B. M. Shahria Alam
*Dept. of Computer Science and Engineering*
East West University
Dhaka, Bangladesh
bmshahria@gmail.com

Md. Nasim Ahmed
*Dept. of Electrical and Computer Engineering*
North South University
Dhaka, Bangladesh
nasimshaikat2702 @gmail.com

*Abstract*— Plant disease diagnosis is essential to farmers' management choices because plant diseases frequently lower crop yield and product quality. For harvests to flourish and agricultural productivity to boost, grape leaf disease detection is important. The plant disease dataset contains grape leaf diseases total of 9,032 images of four classes, among them three classes are leaf diseases, and the other one is healthy leaves. After rigorous pre-processing dataset was split (70% training, 20% validation, 10% testing), and two pre-trained models were deployed: InceptionV3 and Xception. Xception shows a promising result of 96.23% accuracy, which is remarkable than InceptionV3. Adversarial Training is used for robustness, along with more transparency. Grad-CAM is integrated to confirm the leaf disease. Finally deployed a web application using Streamlit with a heatmap visualization and prediction with confidence level for robust grape leaf disease classification.

## I. Introduction

Grape is a delicious fruit. There are several products which is obtained from grapes. Early disease detection becomes essential for economically significant fruits, and the amplification of disease greatly raises losses [1]. The study scope encompassed the use of most current AI and feature reduction techniques for the early detection and classification of grape leaf disease.

Combine of deep features from CNNs and SVMs helps efficient grape leaf disease detection [2]. Various study with CNN designs with deep feature layers and fusion approaches helps to improve the accuracy of grape leaf disease recognition. One of China's significant fruit industries is the grape industry. Through early disease detection in the leaf, artificial intelligence will assist them in diversifying the market and boost grape production [3].

Grape's cultivation makes a huge effect on the economy and content of diverse societies around the world. Both Deep learning and machine learning methods are popular options for image processing, particularly for the detection of plant leaf disease detection [4]. Additionally, this study provides solutions to enhance the accuracy and effectiveness of recognizing grape leaf disease, which could have an immense impact on the agriculture sector.

In this context, our research was guided by two primary questions: (1) Is it possible for deep learning architectures to concurrently achieve high accuracy and interpretability in the classification grape leaf disease? (2) Does the integration of explainability and robustness testing offer a viable approach to scalable real-time grape leaf disease classification?

To address these questions, we introduced LeafLife, an explainable deep learning framework designed for automated classification and severity grading grape leaf disease. Our contributions are threefold. First, we evaluated three state-of-the-art convolutional neural network (CNN) architectures, InceptionV3, Xception. Second, we incorporated explainable artificial intelligence (XAI) techniques, including Gradient-weighted Class Activation Mapping (Grad-CAM) to provide visual rationales that highlight the spots. Third, we evaluated the robustness of the most effective model under adversarial perturbations and demonstrated a lightweight web-based prototype to illustrate the potential of the workflow.

The remainder of this paper is organized as follows. Section II provides a review of related. Section III outlines the dataset, and methodology. Section IV presents the experimental setup and results. Section V discusses the findings and robustness. Section VI presents a comparison of the proposed framework with existing works. Section VII concludes the study with a summary of the contributions, identified limitations, and suggestions for future research.

## II. Related Works

The DexNet framework deploys nine domain-adapted CNN architecture as critical to deal with the problem of limited training data in plant disease [5]. The DexNet achieved 89.06%, 92.46%, abd 94.07% accuracy for 5-shot, 10-shot, and 15-shot classification tasks on PlantVillage dataset. However, there is a drawback in the environments with limited resources.

DCNN classifier is a based architecture for plant leaf classification [6]. Qin et al. [7] uses different techniques among them the SVM model is noticeable because of the it achieved 94.74% accuracy. In the study some data was unlabelled so they proposed a semi-supervise method for getter better result on the dataset.

The present research an automated system that employs machine learning and image processing in grape leaf disease detection [8]. Using the GrabCut method they achieved 93% accuracy on testing by SVM model. Depends on the manual segmentation and possible sensitivity to shifts in image quality are among the limitations.

In order to improve obtain feature for grape leaf disease detection. In the study [9] they proposed CNN based architecture and United Model that combines multiple CNNs. It can distinguish between leaves that are healthy and those that have esca, isariopsis leaf spot, and black rot after being trained on the PlantVillage dataset. Potential overfitting and the requirement for a diverse dataset to enhance generalization are among the limitations.

Javidan et al. [10] achieved 94% accuracy using K-means clustering on a grape leaf disease dataset which contains 3.885 images along with four classes. Among them three are leaf diseases and other one is healthy leaf. After using K-means clustering they applied SVM multi-class classification.

Despite these advances, three challenges remain, namely limited interpretability, insufficient robustness testing, and weak clinical integration. This study addressed these gaps by benchmarking CNN architectures, incorporating interpretability through Grad-CAM and occlusion sensitivity, assessing robustness through adversarial perturbations, and developing a prototype web application..

## III. METHODOLOGY

Fig. 1 illustrates the methodological framework of this study. It overall shows the overview of this study. The process commences with the preparation and preprocessing of the dataset, followed by the development of models across two convolutional neural network (CNN) architectures. Also focused on adversarial robustness. A proto type web application also demonstrate in the study.

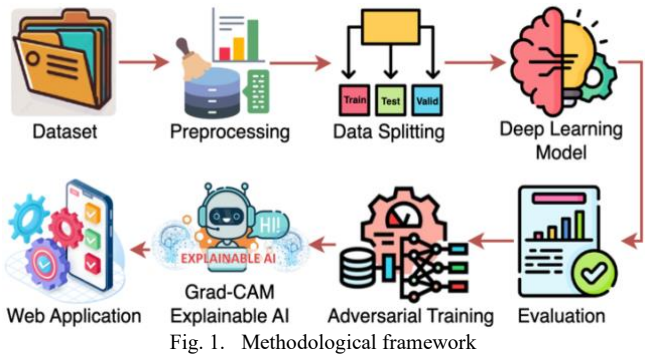

Fig. 1. Methodological framework

### A. Dataset

In this study the used dataset contains 9,032 images from Kaggle [11]. This dataset has four classes: Black Rot (2360), Black Measles (2400), Isariopsis Leaf Spot (2152), and Healthy (2115). There were more Black Measles and Black Rot images than other images. This imbalance makes it difficult for the models to work well. Fig. 2 shows examples of images from each class.

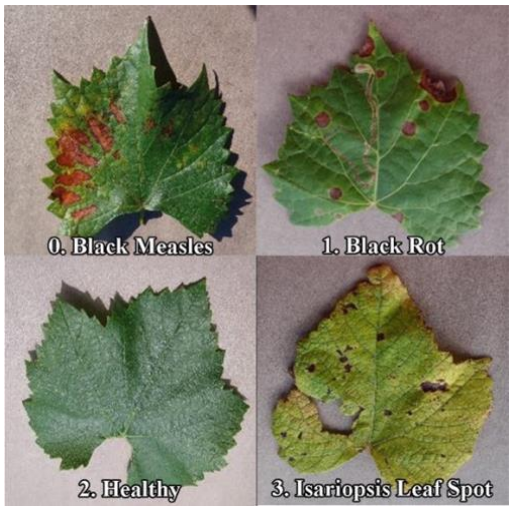

Fig. 2. Sample images from dataset

### B. Preprocessing

All images were resized to 224×224 pixels to align with the input specifications of the pre-trained models to facilitate stable optimization. The dataset was divided into subsets comprising 70% training, 20% validation, and 10% testing, ensuring that the evaluation was performed on previously unseen samples.

### C. Model Architectures

We benchmarked two different CNN architectures for the study.
- *InceptionV3* employs inception modules to capture features at multiple scales while reducing the computational costs.
- *Xception* is an extended form of inception with depthwise separable convolutions to improve efficiency.

These architectures were selected These architectures were selected for their proven success in medical imaging tasks and complementary design characteristics

### D. Training Setup

Two models with same training method was used. The Adam optimizer has a learning rate of 0.0001. For multiple classes suitable loss function Categorical cross-entropy was used. The batch size was set as 32. To prevent overfitting, early stopping was used with patience for five epochs as shown in Table I.

TABLE I. HYPERPARAMETER SETTINGS FOR THE EVALUATED CNN ARCHITECTURES

| Parameter | InceptionV3 | Xception |
|---|---|---|
| Batch size | 32 | 32 |
| Loss function | Categorical cross-entropy | Categorical cross-entropy |
| Learning rate | 0.0001 | 0.0001 |
| Optimizer | Adam | Adam |
| Number of epochs | 45 | 54 |
| Early stopping (patience) | 5 | 5 |

## IV. RESULTS

This section evaluates the performances of the two CNN architectures used in this study.

### A. InceptionV3

Fig. 3 illustrates a sharp decline in training loss reaching about 0.2 while training accuracy peaks at 90%. About 80% the validation accuracy can be reached for the InceptionV3.

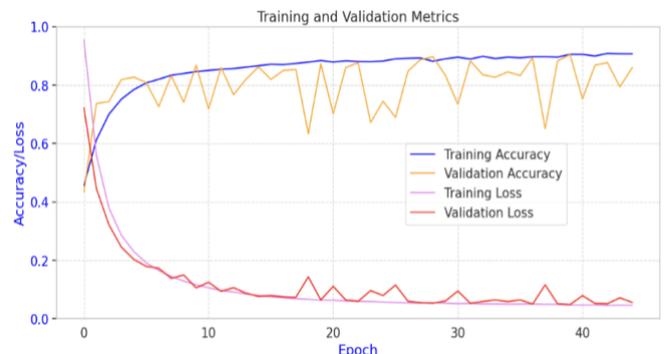

Fig. 3. Loss for training and validation of InceptionV3 architecture

The confusion matrix presented in Fig. 4 reveals high accuracy along with some misclassifications between classes. Notable between the Black Rot and Black Measles. Although Healthy was correctly predicted 217 times and Isariopsis Leaf Spot was correctly predicted 207 times.

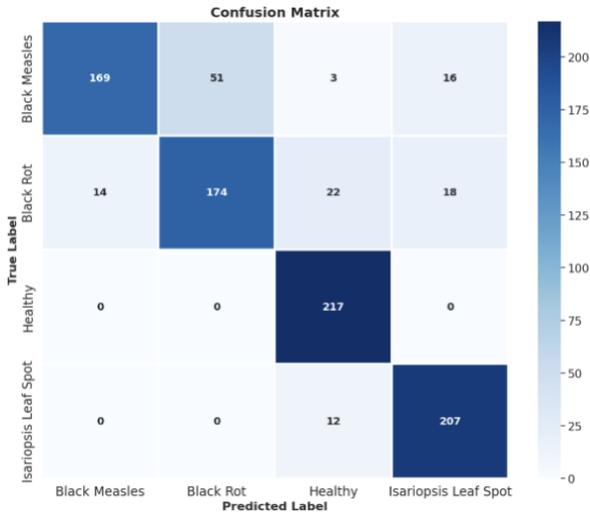

Fig. 4. Confusion Matrix of InceptionV3 architecture

*B. Xception*

Fig. 5 illustrates the performance of Xception, with the training and validation curves demonstrating strong convergence and validation accuracy consistently exceeding 97%.

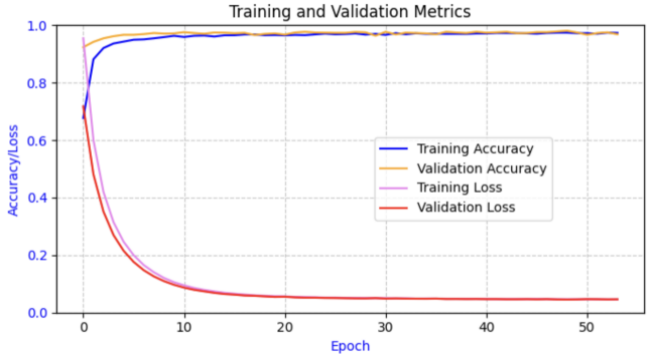

Fig. 5. Training and validation accuracy and loss curve of Xception archirecture

The confusion matrix depicted in Fig. 6 indicates minor misclassification comparison to InceptionV3. Xception shows mostly correct prediction across the classes. It almost shows perfect prediction for Isariopsis Leaf Spot and Healthy.

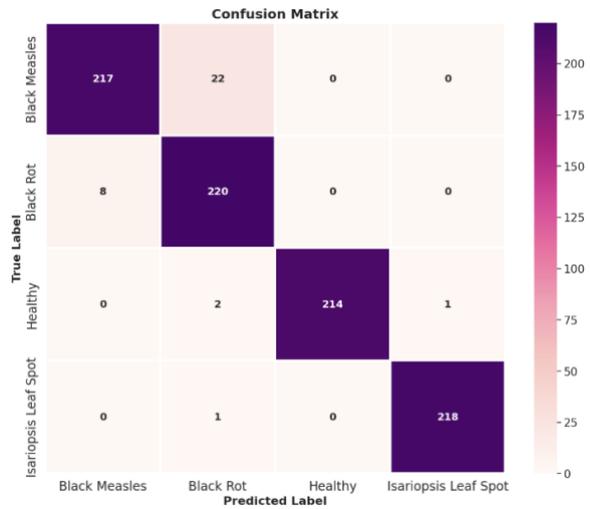

Fig. 6. Confusion Matrix of Xception architecture

*C. Adversarial Robustness*

As adversarial training represent robustness of the model. Here our proposed Xception achieved 98.09% peak accuracy at ε = 0.1. Table II demonstrate the robustness summary of the Xception architecture with works promising with noisy images.

TABLE II. ADVERSARIAL ROBUSTNESS EVALUATION OF XCEPTION UNDER VARYING PERTURBATION LEVELS ($\varepsilon$)

| Epsilon Value ($\varepsilon$) | Validation Loss | Validation Accuracy | Optimal Epochs |
|---|---|---|---|
| 0 | 0.1645 | 0.9766 | 3 |
| 0.1 | 0.1448 | 0.9797 | 13 |
| 0.12 | 0.1476 | 0.9797 | 5 |
| 0.14 | 0.1450 | 0.9803 | 4 |
| 0.16 | 0.1412 | 0.9809 | 13 |
| 0.18 | 0.1405 | 0.9784 | 8 |
| 0.2 | 0.1381 | 0.9784 | 24 |

## V. EXPLAINABLE AI

In addition to being accurate and reliable, models must be easy to understand for use in LeafLife applications.

*A. Grad-CAM Visualizations*

Fig. 7 illustrate the Grad-CAM where it highlighted all the spots in the leaf. It leads the application to determine which part of the leaf is affected for better prediction.

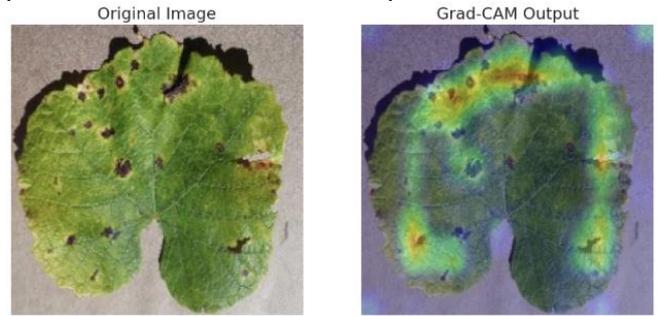

Fig. 7. Grape leaf disease identification using Grad-CAM

## VI. PROTOTYPE WEB APPLICATION

To illustrate the practical applicability of the proposed framework, we developed a lightweight web-based application called LeafLife shown in Fig. 8.

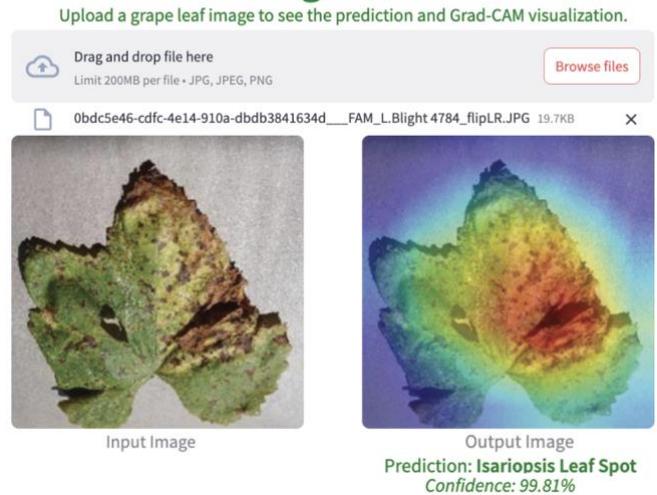

Fig. 8. LeafLife web application for grape leaf disease classification

The application is based on the Xecption models training.

- Predicted classes (Black Measles, Black Rot, Healthy, Isariopsis Leaf Spot).

- A Grad-CAM heatmap overlay highlights the spots for better identification and confidence score qualifies the prediction certainty.

Although LeafLife is yet to undergo usability testing or validation in real-world environments, its design exemplifies the potential for delivering explainable deep learning models in an accessible format.

## VII. COMPARISON AND DISCUSSION

To situate the performance of the proposed LeafLife framework within the existing body of research, we conducted a comparative analysis with the two CNN architectures for grape leaf disease classification.

### A. Evaluation of CNN Architectures

Table III presents a comparative analysis of the performance metrics for InceptionV3 and Xception. Xception achieved highest 96.23% test accuracy between the, which had accuracies of 84.94% and 96.23% respectively. According to the classification report Xception shows noticeable values than InceptionV3.

TABLE III. PERFORMANCE COMPARISON OF EVALUATED CNN ARCHITECTURES

| Model Name | Classes | Precision | Recall | F1-Score |
|---|---|---|---|---|
| InceptionV3 | Black Measles | 92.35 | 70.71 | 80.09 |
| | Black Rot | 77.33 | 76.82 | 76.82 |
| | Healthy | 85.43 | 1.00 | 92.14 |
| | Isariopsis Leaf Spot | 85.89 | 94.52 | 90.00 |
| Xception | Black Measles | 96.44 | 90.79 | 93.53 |
| | Black Rot | 89.80 | 96.49 | 93.02 |
| | Healthy | 1.00 | 98.62 | 99.30 |
| | Isariopsis Leaf Spot | 99.54 | 99.54 | 99.54 |

### B. Benchmarking Against Prior Work

We conducted a comparative analysis of LeafLife and existing studies on grape leaf disease detection. As illustrated in Table IV, the Xception architecture achieved an accuracy of 96.23%, surpassing that of several previous methodologies. For instance, Ahmed et al. [5] reported an accuracy of 94.07% using DexNet on PlantVillage dataset, whereas Jaisakthi et al. [8] achieved an accuracy of 93% using SVM. Javidan et al. [10] achieved 94% accuracy K-mean clustering on SVM multi-class classification; however, their performance was compromised by the redundant features. In contrast, LeafLife consistently demonstrated superior accuracy and minimized class-level misclassification. LeafLife contributes to the advancement of this field by incorporating Grad-CAM visualizations and adversarial robustness analysis.

TABLE IV. COMPARISON OF LEAFLIFE WITH PREVIOUS WORKS ON GRAP LRAF DISEASE CLASSIFICATION

| Author | Method | Accuracy (%) |
|---|---|---|
| Ahmed et al. [5] | DexNet | 94.07 |
| Jaisakthi et al. [8] | SVM | 93 |
| Javidan et al. [10] | K-means clustering | 94 |
| Proposed Model | InceptionV3 | 84.94 |
| | Xception | 96.23 |

### C. Discussion of Limitations and Future Work

This comparative analysis provides three principal contributions. Xception exhibited the most reliable classification performance among the evaluated architectures. Second, LeafLife exceeds several documented approaches in terms of accuracy while also integrating interpretability and robustness. Third, Grad-CAM analyses verified that the model depends on spots on leaf, which enhancing trust.

This study had several limitations. The dataset contains only four classes which were taken in controlled lighting condition. Although adversarial training tests confirmed the stability against noise, which helps to shifts across the imaging centers and devices. Future research should focus on enhancing dataset diversity from real environment.

## VIII. CONCLUSION

This study introduced LeafLife, a deep learning tool that helps classify the grape leaf disease. It tested two types of neural networks InceptionV3, and Xception on the dataset. Xception performed the best with a test accuracy of 96.23% and was robust against small changes in the images. Grad-CAM tests were used to simplify the interpretation of the results. A web application was also created to demonstrate how LeafLife can be used as a tool to help doctors make decisions.

Although the results of this study are promising, several limitations remain. The dataset was collected in optimum condition. These factors may affect to the results. Despite these challenges, LeafLife demonstrates that integrating accuracy, robustness, and clarity is an effective approach to utilizing AI to assess grape leaf disease detection.

REFERENCES

[1] Atesoglu, F., & Bingol, H. (2025). The Detection and Classification of Grape Leaf Diseases with an Improved Hybrid Model Based on Feature Engineering and AI. *AgriEngineering*, *7*(7), 228. https://doi.org/10.3390/agriengineering7070228

[2] Patil, R. G., & More, A. (2024). Grape leaf disease diagnosis system using fused deep learning features based system. Procedia Computer Science, 235, 372–382. https://doi.org/10.1016/j.procs.2024.04.037

[3] Liu, B., Ding, Z., Tian, L., He, D., Li, S., & Wang, H. (2020). Grape leaf disease identification using improved deep convolutional neural networks. *Frontiers in Plant Science, 11*, Article 1082. https://doi.org/10.3389/fpls.2020.01082

[4] Aher, P. G., Sabnis, V., & Jain, J. K. (2025). Deep learning for grape leaf disease detection: A review . *Multidisciplinary Reviews*, *8*(11), 2025364. https://doi.org/10.31893/multirev.2025364

[5] Ahmed, S., Hasan, M. B., Ahmed, T., & Kabir, M. H. (2025). DExNet: Combining observations of domain adapted critics for leaf disease classification with limited data. Scientific Reports, 15, 9002. https://doi.org/10.1038/s41598-025-59562-x

[6] Prasad, K. V., Vaidya, H., Rajashekhar, C., Karekal, K. S., Sali, R., & Nisar, K. S. (2024). Multiclass classification of diseased grape leaf identification using deep convolutional neural network (DCNN) classifier. *Scientific Reports, 14*, 9002.

[7] Qin, F., Liu, D., Sun, B., Ruan, L., Ma, Z., & Wang, H. (2016). Identification of alfalfa leaf diseases using image recognition technology. *PLOS ONE, 11*(12), e0168274. https://doi.org/10.1371/journal.pone.0168274

[8] Jaisakthi, S. M., Mirunalini, P., & Thenmozhi, D. (2019). Grape leaf disease identification using machine learning techniques. In *Proceedings of the 2019 International Conference on Computational Intelligence in Data Science (ICCIDS)* (pp. 1–6). IEEE. https://doi.org/10.1109/ICCIDS.2019.8862084

[9] Ghuge, A., Jagtap, D., Sangle, S., Darade, D., & Rumale, A. (2024). Grape leaf disease detection using image processing and CNN. *International Journal of Advanced Research in Computer and Communication Engineering*, 13(4), 1181–1186. https://doi.org/10.17148/IJARCCE.2024.134176

[10] Javidan, S. M., Banakar, A., Asefpour Vakilian, K., & Ampatzidis, Y. (2022). Diagnosis of grape leaf diseases using automatic K-means clustering and machine learning. *SSRN Electronic Journal*. https://doi.org/10.2139/ssrn.4062708

[11] Vipooool. (n.d.). *New Plant Diseases Dataset*. Kaggle. https://www.kaggle.com/datasets/vipoooool/new-plant-diseases-dataset